\documentclass{article}
\usepackage[english]{babel}
\usepackage[square,numbers]{natbib}
\usepackage[pdftex]{graphicx}
\usepackage[capitalize]{cleveref}
\usepackage{authblk}

\title{RandomSCM: interpretable ensembles of\\sparse classifiers tailored for omics data}

\author[1,2]{Thibaud Godon}
\author[2]{Pier-Luc Plante}
\author[1]{Baptiste Bauvin}
\author[1,2]{Élina Francovic-Fontaine} 
\author[3]{Alexandre Drouin}
\author[1,2]{Jacques Corbeil}

\affil[1]{Department of Computer Science and Software, Universit\'e Laval}
\affil[2]{Department of Molecular Medicine, Universit\'e Laval}
\affil[3]{Element AI, a ServiceNow company}
\date {}

\begin{document}

\maketitle

Recent metabolomics measurement instrumentation, such as mass spectrometers, produce extremely high-dimensional data. This situation is in sharp contrast with the small sample sizes that are typically encountered in this setting. Together, these make for conditions that challenge most statistical methods~\citep{Gromski2015}, which are known as the fat data (or $p \gg n$) problem. Machine learning (ML) algorithms that rely on sparsity to predict phenotypes using very few covariates have been shown to thrive in this setting. While sparsity helps to avoid overfitting, it also leads to concise models that are easier to interpret for biomarker discovery.

The Set Covering Machine (SCM) algorithm \citep{scm} is one such method. It has the particularity of producing sparse models that can be interpreted as simple decision rules. Recent work has applied SCMs to the genotype-to-phenotype prediction of antibiotic resistance and achieved state-of-the-art accuracy \citep{drouin2019interpretable}. While SCMs were shown to work well for simple phenotypes, the models that they learn are limited to a single conjunction or disjunction of rules. Our work aims to relax this limitation while keeping the models concise and easy to interpret. We start from the observation that SCM models can be interpreted as special cases of decision trees \citep{drouin2019interpretable} and follow the path of Random Forests \citep{randomforest}. That is, we propose RandomSCM: a bootstrap aggregation of SCM models.

\begin{figure}
\centering
\includegraphics[width=320pt]{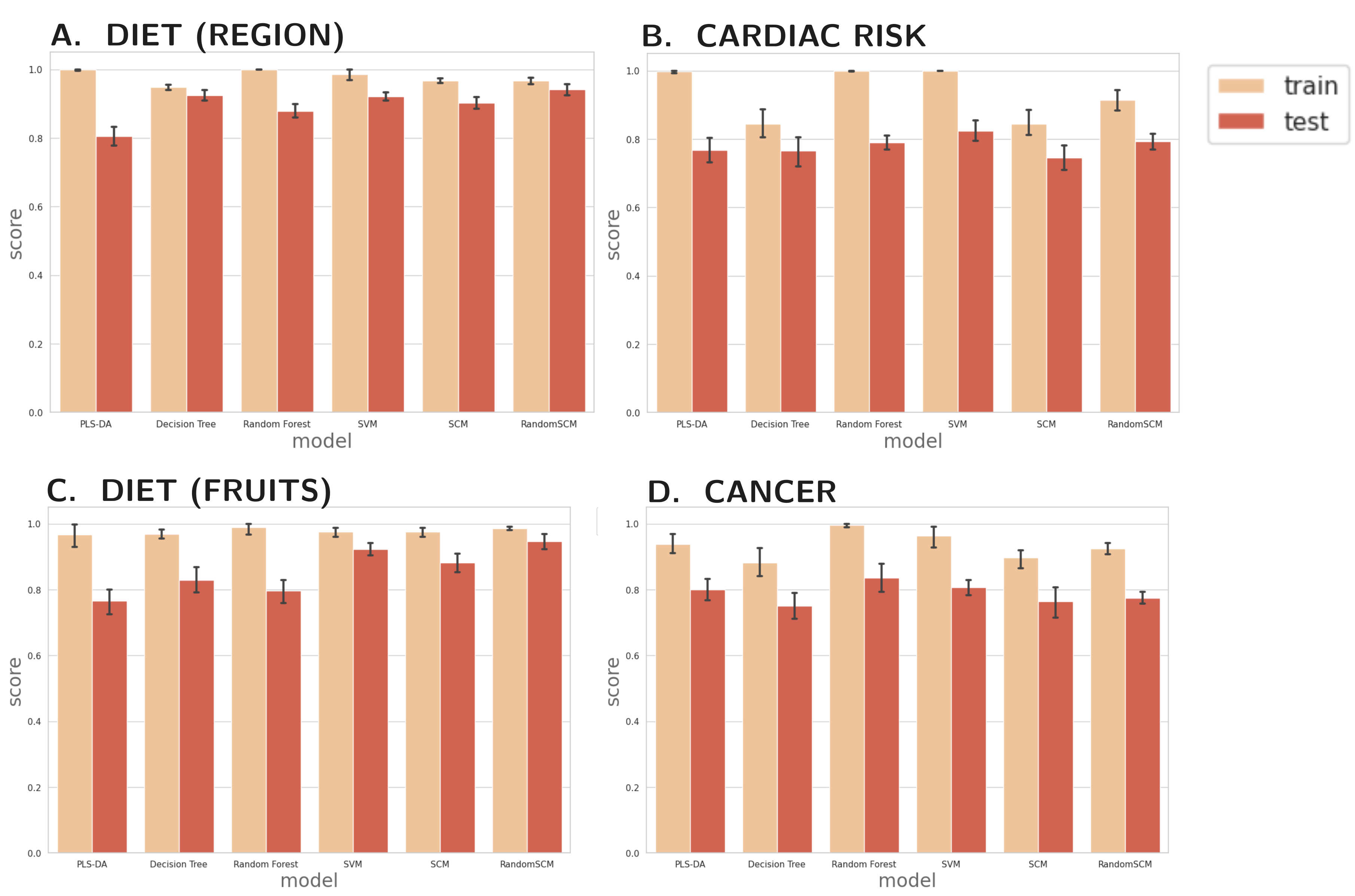}
\caption{RandomSCM (last column) prediction scores compared to state-of-the-art machine learning algorithms, on 4 metabolomics datasets.}
\label{fig:results}
\end{figure}

We explore applications of RandomSCM beyond genotype-to-phenotype prediction by applying it to four metabolomics datasets, where the input consists of liquid chromatography mass spectometry (LC-MS) data and the output is a binary label.
The performance of RandomSCM is compared to common statistical methods and learning algorithms, namely: partial least square discriminant analysis (PLS-DA), decision trees~\citep{decisiontree}, random forests~\citep{randomforest}, support vector machines (SVM)~\citep{svm1995}, and SCM.
Our results, shown in \cref{fig:results}, demonstrate the state-of-the-art performance of RandomSCM on different public and in-house metabolomics datasets.
We make two important observations. First, RandomSCM ensembles are strictly more accurate than SCM models, indicating that there are clear benefits to considering models slightly more expressive than those of SCM in the context of metabolomics data.
Second, the gaps between the training and testing errors of RandomSCM are always smaller than those of random forests. This suggests that relying on simple decision rules such as conjunctions and disjunctions, instead of decision trees, reduces overfitting in the fat data setting, in line with observations of \citep{drouin2019interpretable}.

\begin{figure}
\centering
\includegraphics[width=300pt]{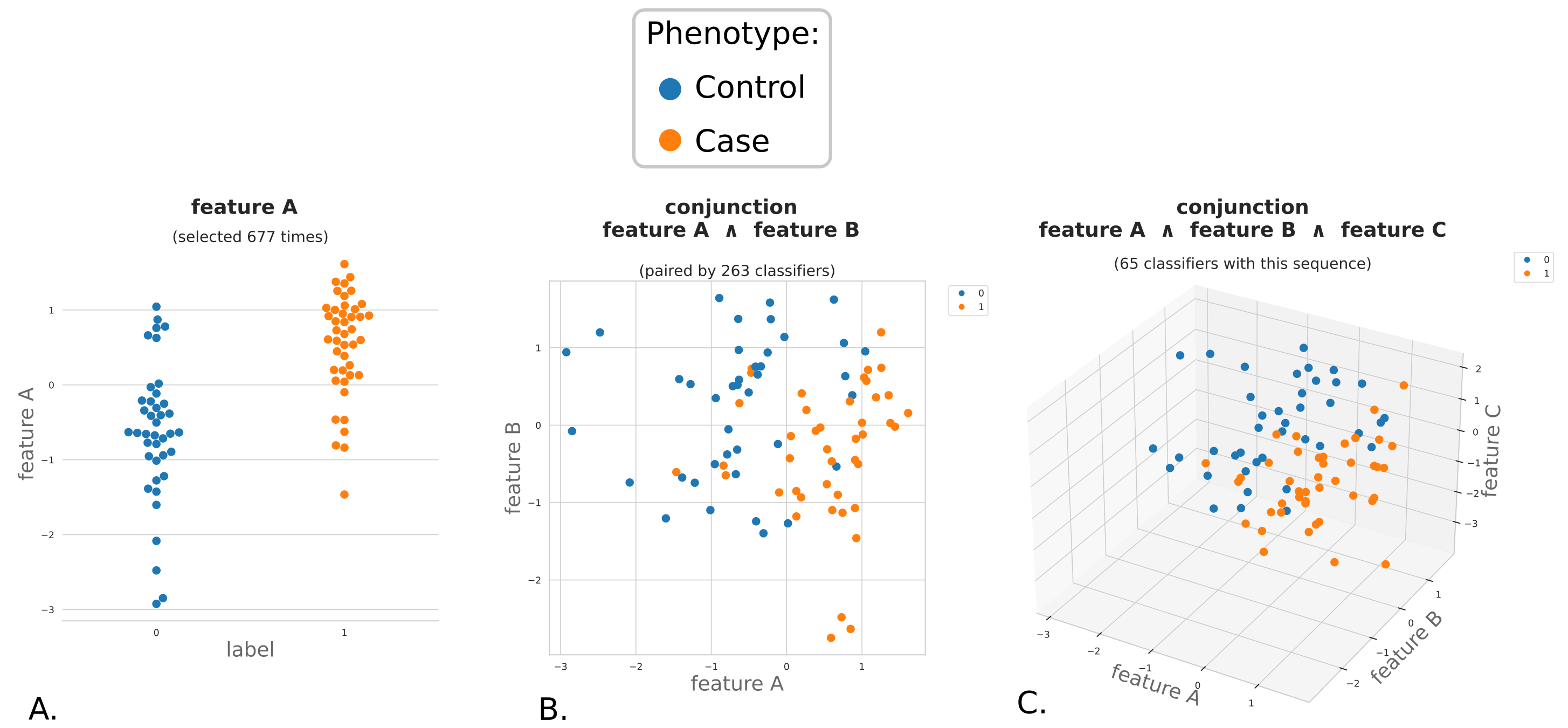}
\caption{Projections of a dataset on the most frequents conjunctions of features in a RandomSCM estimator fitted to predict the phenotype (patients with Gatric Cancer). This figure is used to find pathways that are predictive of the biological state. a) Most frequent feature (conjunction of size 1), b) most frequent pair of features, c) most frequent conjunction of 3 features.}
\label{fig:dimensionplots}
\end{figure}

Furthermore, a study of the decision rules in the fitted RandomSCM models revealed biologically relevant biomarkers.
For each ensemble, we analysed the most common conjunctions of rules and how well they separated the classes.
\cref{fig:dimensionplots} illustrates this for the most common conjunctions of size one (\cref{fig:dimensionplots} a), two (\cref{fig:dimensionplots} b), and three (\cref{fig:dimensionplots} c) in the \texttt{gastric-cancer} \citep{metaboref} dataset (corresponding to \cref{fig:results} D).
In this case, the RandomSCM model recovered 11 of 22 biomarkers that had previously been reported in the literature.
Similar results were obtained for the \texttt{mediterranean-diet} dataset, where the most common features used in the conjunctions correspond to characteristic compounds present in the Mediterranean diet, which were correctly identified out of 15089 putative metabolites.

Overall, these results demonstrate the high potential of the RandomSCM algorithm for biomarker discovery in high-dimensional biological data, which abounds in metabolomics studies.
In future work, we aim to extend the applicability of this method to other types of omics data, where the number of features is again much greater than the number of examples (e.g., proteomics).

\medskip
\bibliography{main}

\end{document}